# Exploring the Role of Logically Related Non-Question Phrases for Answering Why-Questions

Niraj Kumar[1], Rashmi Gangadharia[2], Kannan Srinathan[3], Vasudeva Varma[4],

niraj_kumar@research.iiit.ac.in[1], rashgang@in.ibm.com[2], srinathan@iiit.ac.in[3], vv@iiit.ac.in[4],

{IIIT-Hyderabad, Hyderabad-500032}[1,3,4], {IBM Research}[2], INDIA

**Abstract.** In this paper, we show that certain phrases although not present in a given question/query, play a very important role in answering the question. Exploring the role of such phrases in answering questions not only reduces the dependency on matching question phrases for extracting answers, but also improves the quality of the extracted answers. Here matching question phrases means phrases which co-occur in given question and candidate answers. To achieve the above discussed goal, we introduce a bigram-based word graph model populated with semantic and topical relatedness of terms in the given document. Next, we apply an improved version of ranking with a prior-based approach, which ranks all words in the candidate document with respect to a set of root words (i.e. non-stopwords present in the question and in the candidate document). As a result, terms logically related to the root words are scored higher than terms that are not related to the root words. Experimental results show that our devised system performs better than state-of-the-art for the task of answering *Why*-questions.

## 1 Introduction

According to [5], about 5% of all questions asked in QA systems are *why*-questions. *Why*-questions have been less explored when compared to factoid-based question answering tasks.

Answers to these questions contain a reason or a cause and in majority of the cases, the answers do not have a high lexical similarity with the question. Even linguistic features fail to identify the answers correctly. The task of answering *why*-questions has been treated as a difficult task due to the low performance of various systems suggested in the past [14]. Additionally, due to variability in the length of the answer, higher difficulty level and pragmatic nature, the task of *Why*-based question answering becomes a challenging task of significant interest.

To get an effective solution, the task is devised as a two step process. The first step involves identification of a few candidate/relevant documents that most likely contain the answer. The next step generates a ranked list of answers extracted from these relevant documents (named as Answer Extraction). It is important to note that the technique suggested in this paper can be easily extended to community-based answer ranking with some additional external features and clues. The contribution of this paper can be summarized as follows:

- We introduce a bigram-based word graph model of text, which captures semantic and topical information of text. It also capture the information of overlapping terms of given question (i.e., which co-occur in given question and candidate document). The main aim of this step is to utilize this information in answering the given question.
- We finally introduce a modified scheme for ranking with priors, which makes use of the information captured in text graph. This arrangement gives high scores to both types of terms, i.e. (1) terms which exist in the given question and (2) terms which are not present in the given question but are logically important for answering the question.

**Brief Description of System:** First of all, we convert the given question into query, which contains sequence of non-stopwords terms connected by appropriate "AND" and "OR" relational operators. We pass this query to Solr[1] and extract top 100 documents. We also note their relevance score (See Section 4). Next, we rank candidate answers in each extracted document (See Section 5). Finally, to get the final ranked list of answers, we combine the document's relevance sore (w.r.t. given question) and rank score of candidate answer (See Section 6).

## 2 Problem Definition and Motivation

### 2.1 Problem Definition

Majority of times, matching and/or overlapping phrases/terms, which are common in given question and candidate answers may not give the accurate answer. To understand this problem, we go through a simple example, which contains a sample question; human annotator's selected answer and candidate answers (See Table 1).

From Table 1, it is clear that the most relevant answer as selected by the human annotators and paragraphs, "PID-3" and "PID-4" contain the same number of matching non-stopword question terms, like: "white", "flag", "symbol" and "surrender".

This doesn't mean that, we are neglecting the importance of non-stopword terms which are common to both, i.e., given question and candidate answers, in extracting answer(s) for any given question. But, in addition to the importance of such terms, we are trying to explore the role of some other terms that are logically related to the terms in the question but not present in question.

---

[1] http://lucene.apache.org/solr/

If we go through the given/selected answer, then, we find that there are some other terms like: "negotiation", "internationally", "fired", "signifies", "unarmed", "waving", "person", "geneva", "convention", etc. (see, Table 1, bold and underlined words in given answer), plays very important role in answer, but not present in question. So, now the main problem is how to identify/give relatively higher rank to such terms. It is important to note that, identifying such kind of logically related terms by using only semantic relatedness or linguistic features is very tough.

Table 1: Sample Question with selected answer and other candidate answers (source: [11], [12])

| |
|---|
| **Question:** Why is a **white flag** a **symbol** of **surrender**? |
| **Given/selected answer by annotator (**Source Document: White_flag.htm**):** The **white flag** is an *internationally recognized* **protective sign** of **truce** or **ceasefire**, and request for *negotiation*. It is also used to *symbolize* **surrender**, since it is often the **weaker military party** which requests *negotiation*. A **white flag** *signifies* to all that an approaching **negotiator** is *unarmed*, with an intent to **surrender** or a *desire* to communicate. *Persons* carrying or *waving* a **white flag** are not to be *fired* upon, nor are they allowed to open *fire*. The use of the **flag** to **surrender** is included in the *Geneva* **Conventions**. |
| **Other candidate answers:** |
| **PID-4 #** The first mention of the usage of **white flag**s to **surrender** is made during from the Eastern Han dynasty (A.D 25-220). In the Roman Empire, the historian Cornelius Tacitus mentions a **white flag** of **surrender** in A.D. 109. Before that time, Roman armies would **surrender** by holding their shields above their heads. The usage of the **white flag** has since spread worldwide. |
| **PID-3 #** Many times since the weaker party is in a decrepit state, a **white flag** would be fashioned out of anything readily available like a T Shirt, handkerchief, anything **white**. The most common way of making a **white flag** is to obtain a pole and tie two corners of a sheet of cloth to the top of the pole and somewhere in the middle. |

## 2.2 Motivation

To achieve the above discussed goal, we propose a bigram based word graph model. To improve the importance of word pairs, which co-occur in given question and candidate answers, we apply simple link boosting. To properly capture the semantic and topical information in document, we populate the graph with semantic and topical relatedness information. Finally, we modify the "ranking with prior" based scheme, to get higher rank for (1) terms which exist in the given question and (2) terms which are not present in the given question but are logically important for answering the question.

**Ranking with prior:** For any graph $G = (v, e)$, where, $v = \{v_1, v_2, v_3, ... v_N\}$ represents the set of vertices and $E = (v_i, v_j)$ represents the edges, if there exist a link between $v_i$ & $v_j$. The ranking with prior score [15] of any node '$v$' of the graph can be given as:

$$PPR(v)^{i+1} = (1-\beta)\left(\sum_{u \in adj(v)} P\left(\frac{v}{u}\right) PPR(v)^i\right) + \beta P_v \qquad (1)$$

Where, $PPR(v)^{i+1}$ represents the page rank with prior of node 'v' at $(i+1)^{th}$ iteration, $adj(v)$ represents the adjacent node of node 'v'. If 'R' represents the set of root nodes then, prior or bias can be given as: $P_v$ and can be explained as $P_v = \begin{cases} 1/|R| & for\ v \in R \\ 0 & otherwise \end{cases}$

$\beta$ represents the back propagation probability $(0 \leq \beta \leq 1)$, determines how often we jump back to node 'v'.

**Description:** According to [15], in ranking with prior, root is considered as data analyst's prior knowledge or bias in terms of which nodes are considered important in a graph. If we select a root set that encompasses the entire graph, the relative importance converges to the graph's importance. Figure 1, presents an example showing the differences between the traditional Page Rank approach ("Page Rank", [7]) and ranking with priors (obtained by considering node 'A' and 'F' as root nodes) by using a toy graph.

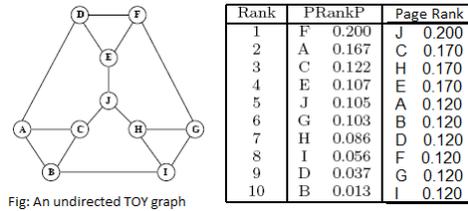

Figure 1: Effect of ranking with prior w.r.t. Root node 'A' and 'F', see score "PRankP", (ref: [15]).

**Why modification in ranking with prior?:** The main aim behind this modification is to avoid the effect of noisy words and to exploit the information useful for answering the question. For this we use bigram model based weighted word graph, which contains (1) additional boosting links, to increase the importance of overlapping phrases from question, (2) semantic importance, to capture the semantic strength of word pairs in text and finally, (3) topical relatedness to join the topically related terms, which are not directly related. We use these weights to modify the equation of page rank with prior. We use all non-stopword terms of given question as root words. To rank the answers, we use top scored terms, obtained from modified form of ranking with prior.

## 3 Related Work

The task of Question Answering has mainly focussed on answering factoid questions, where answers are usually short phrases such as, named entities. Focus has recently moved towards the task of non-factoid question answering, such as, "*why* questions", "*how-to* questions", etc.

[3], rank candidate answer paragraphs for answering *why*-questions in Japanese. It uses Support Vector Machines and features like: content similarity, causal expressions, and causal relations from two annotated corpora and a dictionary were extracted. [6], also presented a supervised technique, which used sentiment analysis and word classes to answer *why*- questions in Japanese.

[14], proposed a three-step model for *why*-QA: (1) a question-processing module that transforms the input question to a query; (2) an off the-shelf retrieval module that retrieves and ranks passages of text that share content with the input query; and (3) a re-ranking module that adapts the scores of the retrieved passages using structural information from the input question and the retrieved passages.

Although significant work has not been done towards answering *why*-questions, a large number of approaches have been suggested in the past for the task of open domain question answering. [1]; [4]; [9] bridge the lexical chasm between the question and the answer using a machine translation model. Such data may not always be available and in order to obtain a reasonable performance, these techniques require large amounts of such data. [10], presents a BOW-based model, which uses statistical weights based on term frequency, document frequency, passage length, and term density. [8], consider the problem of answering definition questions. They use predicate–argument structures (PAS) for improved answer ranking. [13], compared a number of machine learning techniques in their performance for the task of ranking answers that are described by TF-IDF, a set of 36 linguistically-motivated overlap features and a binary label representing their correctness.

## 4 Identifying Candidate Documents

We use Solr to retrieve documents from INEX corpus for every given "WHY" question. For this we use combination of "OR" and "AND" relational operators. We use "AND" and "OR" operator for corresponding "and" and "or" words in question. Next, We replace all other stopwords and punctuation marks of the given question by "OR" relational operator and we put "AND" relational operator between every word pairs, which lays inside the two "OR" operators. This scheme is similar to [6], used to identify the predefined phrase boundary for keyphrases. Next, we separate all highly frequent verbs from sequence of words through "OR" operator. Here, the main aim is to separate these word sequences from word sequences. For this, we use a collection of 1534 frequent verbs from [6].

Finally, we retrieve top-100 documents for each given question and use the extracted documents to generate the ranked list of answers. We also note the relevance score of each such candidate document.

## 5 Extracting Ranked List of Answers

To extract the ranked list of answers from candidate document (s), we apply modified version of Page Rank with prior on undirected word graph of sentences.

### 5.1 Preparing Word Graph of Sentences

We treat every word of a given document as a node in the graph. We add links between two words, if they co-occur together within a window of size two words (i.e., bigram) in the sentences of the candidate document. Formally, we can define an undirected graph as, $G = (V, E)$, where, $V = \{V_1, V_2, V_3, ... V_N\}$ represents the set of vertices and $E = (V_i, V_j)$ represents the edges, if there exist a link between $V_i$ & $V_j$.

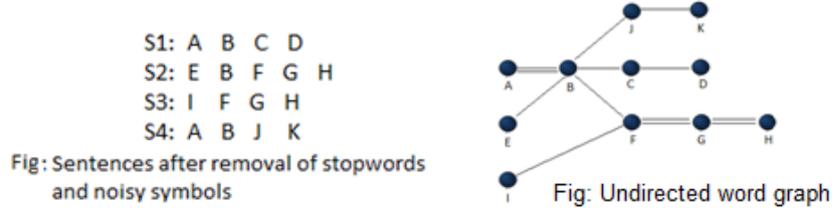

Figure 2: Undirected word graph of sentences, Here S1, S2 and S3 represents the sentences of document and 'A', 'B', 'C', 'D', 'E', 'F', 'G' and 'H' represents the distinct words.

### 5.2 Boosting the Overlapping Phrases

We boost the multi-word overlapping phrases in word graph of sentences, which appear both in the given question and in the source document. For this we add new links on word graph of sentences based on the number of times the matching bigram appeared in the question.

**E.g.,** Suppose a document contains the bigrams "a b", "a d", "b c" and "b e" (Figure 3). Now suppose the question contains a matching bigram "a b". For this we add a new link between "a" and "b". Now, the normalized random walk weight between "a" and "b" is 2/3 (effect of boosting), whereas, the weight between "a" and "d" is 1/3.

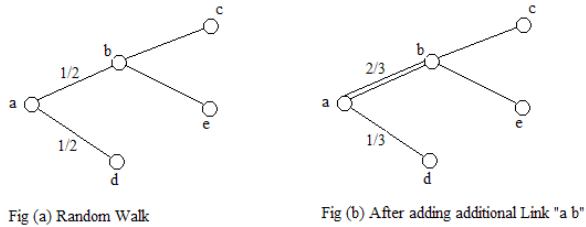

Figure 3: Giving additional boosts to matching phrases

### 5.3 Semantic and Topical Information

We use semantic relatedness score of words of bigrams, to calculate the edge weight of word graph of sentences. Next, we use topical relatedness to identify the topical relation between bigrams if they are not adjacent/directly linked to each other. Finally, we add link between such bigrams and assign weight calculated on the basis of topical relatedness of bigrams.

**Calculating edge weight in word graph by using semantic relatedness score:** For this, we use Wikipedia's extended abstracts[2] and consider only frequent bigrams. This step reduces the participation of noisy/less-important bigrams in calculating the semantic relatedness score. Finally, we use Pointwise Mutual Information to calculate the semantic relatedness strengths of bigrams.

---
[2] http://wiki.dbpedia.org/Downloads38#extended-abstracts

$$PMI(t_i, t_j) = \log_2 \frac{N \times CW(t_i, t_j)}{CW(t_i) \times CW(t_j)} \quad (2)$$

Where, $CW(t_i, t_j)$=Number of Wikipedia extended abstracts which contains terms $t_i$ and $t_j$ in adjacency and having a co-occurrence frequency of at least two. $CW(t_i)$ =Number of Wikipedia extended abstracts, in which $t_i$ occurs at least twice. $N$= total number of Wikipedia extended abstracts.

*Calculating Edge Weight:* Weight of edge between $V_i$ and $V_j$ '$EdgeW(V_i, V_j)$' can be given as:

$$EdgeW(V_i, V_j) = \#links \times PMI(V_i, V_j) \quad (3)$$

Where, $\#links$=Number of links between $V_i$ and $V_j$ (also include additional links after boosting if any such link exists).

**Utilizing topical relatedness score of bigrams:** Co-occurrence of words within a relatively large window in the text suggests that both words are related to the general topic discussed in the text. [16], used such scheme on word-sense disambiguation. We use this concept in identification of topical relatedness of non adjacent bigrams. For this, we use Wikipedia extended abstracts, which contain long abstracts of Wikipedia articles with minor topical deviation from corresponding document-title. However, to reduce the participation of noisy bigrams, we consider only frequent bigrams. Now, we use Pointwise Mutual Information to calculate the topical relatedness between two bigrams:

$$PMI(B_1, B_2) = \log_2 \frac{N \times CW(B_1, B_2)}{CW(B_1) \times CW(B_2)} \quad (4)$$

Where, $B_1$, $B_2$ represents two bigrams. $CW(B_1, B_2)$= Number of Wikipedia extended abstracts in which both $B_1$ and $B_2$ occurs at least twice. $CW(B_1)$=Number of Wikipedia extended abstracts in which $B_1$ occurs at least twice. $CW(B_2)$=Number of Wikipedia extended abstracts in which $B_2$ occurs at least twice.

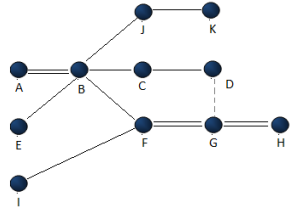

Figure 4: Adding link for topical similarity of bigrams (i.e., "C D" and "G H")

*Adding additional edge:* For bigrams, having semantic relatedness score more than average (calculated by using Eq-4), we add additional edge. The edge weight for such edge can be calculated as:

$$EdgeW(B_1, B_2) = PMI(B_1, B_2) \times Min\{f(B_1), f(B_2)\} \quad (5)$$

Where, $f(B_1)$=Occurrence frequency of $B_1$ in candidate document. $f(B_2)$= Occurrence frequency of $B_2$ in candidate document.

For example (in Figure 4), we add an additional edge between the two topically related bigrams "C D" and "G H" (i.e. between second word of bigram "C D" and first word of bigram "G H"). Here bigram "C D" comes earlier than "G H" in parent document. This scheme adds link between some highly topically related bigrams in word graph.

### 5.4 Applying Ranking with Prior

For this we use word graph of sentences populated with semantic and topical information. Next, we convert it into a row stochastic matrix, by normalizing the row sums of the corresponding transition matrix to one. Finally, we calculate the prior probability of every root word and apply it with modification proposed in ranking with prior.

**Calculating prior probabilities of root words:** Let 'R' represents the set of root words and $P_{|U|}$ denotes the relative importance ("prior bias"), we attach to node 'U'.

$$P_{|U|} = \begin{cases} 1/|R| & For\ 'U' \in R \\ 0 & Otherwise \end{cases} \quad (6)$$

**Ranking with prior:** The modified scheme for ranking with prior is given below:

$$RR(U) = (1-\beta) \left( \sum_{V \in adj(U)} \frac{EdgeW(U,V)}{\sum_{W \in adj(V)} EdgeW(W,V)} \times RR(V) \right) + \beta P_{|U|} \quad (7)$$

Where, $RR(U)$ = Ranking with prior of node 'U'. $EdgeW(B_1, B_2)$ represents edge weight between 'U' and 'V' (See Eq-3,5). $adj(U)$ represents the set of nodes, which are adjacent to node 'U'. $adj(V)$ represents the set of nodes, which are adjacent to node 'V'. $\beta$ represents the back probability $(0 \leq \beta \leq 1)$. It determines how often we jump back to the set of root nodes in 'R'. We use $\beta$ =0.70 (best performance setup, used in all experiments).

Actually, Eq-7 estimates the relative probability of landing on any particular node. By using this equation, we calculate the rank of all words in the given document.

## 6 Ranking Candidate answers

**Ranking candidate answers in each candidate document:** To calculate the scores of a candidate answer, we add the Page rank with prior scores of all the words in the answer. However, to reduce the chances of lengthy candidate answers getting higher ranks, we just use the words whose rank score is greater than the average rank score of all the words. Generally this contains 20-25% of the top ranked words. If it contains more than 25% of total number of words, then we select top 25% of the ranked words. Now, the local score can be calculated as:

$$Score(P_i)_{D_j} = \sum_{W \in P_i} Score(W) \times Relv\_score(D_j) \qquad (8)$$

Where, $P_i$ represents the candidate answer, $Score(P_i)_{D_j}$ = Score of candidate answer $P_i$ in document $D_j$. $Score(W)$ = Rank (score) of word 'W' in document, $D_j$, whose score is more than the average score of all the words, calculated by using Eq-7. $Relv\_score(D_j)$ represents the relevance score of given document $D_j$ for given question (See Section 4, for details). By using same way we calculate the rank score of all candidate answers in each document.

**Final Ranking of Candidate answers:** For this we sort all candidate answers in descending order according to their rank scores.

## 7  Pseudo-code

**Input:** (1) Question, (2) Text collection (here, Wikipedia INEX corpus is used) and (3) Wikipedia extended abstracts.
**Output:** Ranked list of answers for the given question.
**Algorithm:**
St1. We identify the top-100 candidate documents for given question. (Section 4).
St2. Next, we apply following procedures to rank the candidate answers in every candidate documents (St 3. To St 7.).
St3. Prepare word graph of sentences (Sub-section 5.1).
St4. Apply boosting of co-occurring word pairs in word graph, which co-occur in the given question. (Sub-section 5.2).
St5. Add (1) semantic information of words/nodes of every bigram and (2) topical relatedness score of every bigram with word graph of sentences. (Sub-section 5.3).
St6. Apply ranking with prior to calculate the ranks of all words in the candidate document (Sub-section 5.4).
St7. Use ranking with prior score of words in the given document and calculate the scores of every candidate answer. (Section 6).
St8. Take product of relevance score of candidate document and score of related candidate answer to calculate the final score of all candidate answer and then rank all candidate answers in descending order of their scores (Section 6).

## 8  Evaluation

To evaluate our devised system, we used the dataset[3] prepared by [11], [12] and the Wikipedia INEX corpus[4]. We compare the results of our devised system with

---
[3] http://lands.let.ru.nl/~sverbern/

published results of [14] and other baseline system, evaluated on the same set. The details of the dataset and the evaluation metrics are given below:

**Dataset:** We used the Wikipedia INEX corpus [2] as our dataset (contains more than 600,000 XML documents in English). To evaluate the accuracy of the system, we used the gold standard dataset, manually prepared by [11], [12]. This dataset contains 216 questions for which there exists an answer in the Wikipedia collection. These answers are in 206 different documents. This dataset contains 210 answer fragments and manually annotated RST structures (Carlson et al., 2003), in the rs3 file format. This dataset contains a wide variety of *why-* questions like: (1) Questions having no direct match with the title or the main theme of the documents, (2) Questions for which the correct answer passage does not contain phrases that match the phrases in the question, (3) Long questions like: *"Why do Americans cut their meat then put the knife down and change hands with their fork in contrast to most Europeans who work their fork with the opposite hand from their knife?"* , and (4) very short questions like: *"Why does it snow?", "Why do we laugh?"*.

**Evaluation Metrics:** We use the following two evaluation metrics (same as in [14]).

*Success@n:* the number questions that have at least one answer in the top 'n' results.

*MRR (Mean Reciprocal Rank):* It is the average of the reciprocal ranks of the results for a sample of queries:

$$MRR = \frac{1}{|Q|} \sum_{i=1}^{|Q|} \frac{1}{rank_i} \qquad (9)$$

Where, $|Q|$ = Number of queries and $rank_i$ is the rank of the answer of the $i^{th}$ query.

*Note:* If the system does not retrieve any answers in the list of given 'n' top answers, the system gets an RR=0 for the given question. (This approach is same as applied in [14]).

### 8.1 Extracting Ranked List of Answers

To properly evaluate the techniques applied in our devised system, we use four different setups:

**SETUP-1**: In this setup, we consider paragraphs as candidate answers (same as applied by [14]) and prepared a ranked list of paragraphs as answers. See pseudo code (Section 7) for description of the system.

**SETUP-2**: Here we consider 5 consecutive sentences as a candidate answer. This is similar to the approach adopted by [6]. Rest of the system is same, as described in pseudo code (Section 7).

**SETUP-3**: This setup is similar to "SETUP-1", except that we do not use semantic and topical relatedness measure in preparation of graph (i.e., absence of scheme applied in Sub Section 5.3). Rest of the system is same, as described in pseudo code (Section 7).

**SETUP-4**: This setup is similar to "SETUP-3", except that we do not use topical relatedness measure in preparation of graph (See Sub section 5.3 for related description). Rest of the system is same, as described in pseudo code (Section 7).

---

[4] http://www-connex.lip6.fr/~denoyer/wikipediaXML/

Note: As page rank [7], performs poorer then baselines, so we didn't include it in evaluation result.

**Table 2:** Evaluation results

| System | Success@10 | Success@150 | MRR@10 | MRR@150 |
|---|---|---|---|---|
| SETUP-1 | **68.876** | **85.447** | **42.184** | **42.522** |
| SETUP-2 | 66.603 | 82.627 | 40.791 | 41.119 |
| SETUP-3 | 62.47 | 77.5 | 38.26 | 38.568 |
| SETUP-4 | 66.052 | 81.944 | 40.454 | 40.778 |
| Lemur / Tf_IDF-sliding | 45.00% | 78.50% | N/A | 25.00% |
| [14] | 57.00% | 78.50% | N/A | 34.00% |

**Analyses of result:** based on the results given in Table 2:

  i. In both cases i.e., in SETUP-1 and 2, the performances of the systems are nearly the same. This shows the effectiveness of our devised system in generating answers of different format and length, without affecting the quality of answers.
 ii. SETUP-3 performs poorer than SETUP-1 and 2. This is due to the absence of the semantic and topical relatedness score. However, the results are still better than the state-of-the-art (last two rows in Table 2).
iii. Slightly lower performance of SETUP-4 w.r.t. SETUP-1 shows the impact of absence of the topical relatedness score.
 iv. *Comparison with [14]:* We also compared our results with the approach adopted in [14] and "Lemur/tf-idf" based baselines used by [14] (Table 2). The results in Table 2, show that our devised system performs better than both systems.

**Some issues with questions contain only one root word:** For some questions like:

- ✓ Why does it *snow*? (Source: Snow.htm).
- ✓ Why do we *laugh*? (Source: Laughter.htm)

Our devised systems show poor performance, such types of questions i.e., average MRR@10 = 20%. The poor performance could be attributed to the fact that the ranking with prior based system is required to rank the words in the word-graph with respect to a single root word. There are total '4' such questions are given in the dataset. However, with the increase in non-stopword question words or root words (>=2) performance of system shows stability (i.e., in quality of result).

## 9 Conclusion and Scope

In this paper, we used bigram based word graph of sentences, populated with semantic and topical information with some boosted links. Finally, we presented an improved version of "page rank with prior", to rank words in the graph w.r.t. root words in given question. Improvements in the quality of the results show the effectiveness of this scheme.

This scheme can be extended in some other tasks like: (1) guided summarization task (where prior information is supplied to extract the most suitable summary sentences) and (2) community-based question answering systems (by incorporating community-based features and clues), etc.